\documentclass[runningheads]{llncs}
\usepackage{graphicx}
\usepackage{subfigure}
\usepackage{amssymb}
\usepackage{booktabs}
\usepackage{array}
\usepackage{float}
\usepackage{cite}
\usepackage[colorlinks,,linkcolor=blue,citecolor=blue,urlcolor=blue]{hyperref}
\begin{document}
\title{Adaptive Context Selection for Polyp Segmentation}
\titlerunning{Adaptive Context Selection for Polyp Segmentation}
%

\author{ Ruifei Zhang \inst{1}  \and 
Guanbin Li\inst{1}\thanks{Corresponding author is Guanbin Li~(liguanbin@mail.sysu.edu.cn).} \and 
Zhen Li\inst{2}  \and 
Shuguang Cui\inst{2} \and
Dahong Qian\inst{3} \and
Yizhou Yu\inst{4}
}


%
\authorrunning{Zhang et al.}
%
\institute{
 $^1$School of Data and Computer Science, Sun Yat-sen University, Guangzhou, China\\ $^2$Shenzhen Research Institute of Big Data, The Chinese University of Hong Kong, Shenzhen, Guangdong, China\\ $^3$ Institute of Medical Robotics, Shanghai Jiao Tong University, Shanghai, China\\ $^4$Deepwise AI Lab, Beijing, China}
%
\maketitle              




%
%
%
%
\begin{abstract}
Accurate polyp segmentation is of great significance for the diagnosis and treatment of colorectal cancer. However, it has always been very challenging due to the diverse shape and size of polyp. In recent years, state-of-the-art methods have achieved significant breakthroughs in this task with the help of deep convolutional neural networks. However, few algorithms explicitly consider the impact of the size and shape of the polyp and the complex spatial context on the segmentation performance, which results in the algorithms still being powerless for complex samples. In fact, segmentation of polyps of different sizes relies on different local and global contextual information for regional contrast reasoning. To tackle these issues, we propose an adaptive context selection based encoder-decoder framework which is composed of Local Context Attention (LCA) module, Global Context Module (GCM) and Adaptive Selection Module (ASM). Specifically, LCA modules deliver local context features from encoder layers to decoder layers, enhancing the attention to the hard region which is determined by the prediction map of previous layer. GCM aims to further explore the global context features and send to the decoder layers. ASM is used for adaptive selection and aggregation of context features through channel-wise attention. Our proposed approach is evaluated on the EndoScene and Kvasir-SEG Datasets, and shows outstanding performance compared with other state-of-the-art methods. The code is available at \url{https://github.com/ReaFly/ACSNet}.

\end{abstract}
\section{Introduction}
\begin{figure}[htp]
\includegraphics[width=0.9\textwidth]{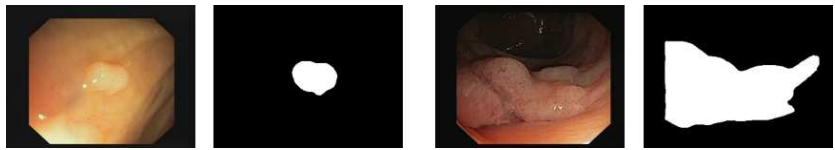}
\centering
\caption{Two examples of polyp segmentation} \label{fig1}
\end{figure}

Colorectal cancer is a serious threat to human health, with the third highest morbidity and mortality among all cancers~\cite{siegel2020cancer}. As one of the most critical precursors of this disease, polyp localization and segmentation play a key role in the early diagnosis and treatment of colorectal cancer. At present, colonoscopy is the most commonly used means of examination, but this process involves manual and thus expensive labor, not to mention its higher misdiagnosis rate~\cite{van2006polyp}. Therefore, automatic and accurate polyp segmentation is of great practical significance. However, polyp segmentation has always been a challenging task due to the diversity of polyp in shape and size. Some examples of polyp segmentation are displayed in Fig.~\ref{fig1}.

In recent years, with the prevalence of deep learning technology, a series of convolutional neural network variants have been applied to polyp segmentation and have made breakthrough progress. Early fully convolutional neural networks~\cite{long2015fully,brandao2017fully,akbari2018polyp,li2018contrast} replaced the fully connected layers of the neural network with convolutional ones. In order to enlarge the receptive field of the neurons, the neural network gradually reduces the scale of the feature map and finally generates the prediction with very low resolution, resulting in a rough segmentation result and prone to inaccurate boundaries. Later, UNet~\cite{ronneberger2015u} based structure was proposed, which adopts a stepwise upsample learning to restore the feature map resolution while maintaining the relatively large receptive field of the neurons. At the same time, the skip connection is used to enhance the fusion of shallow and deep features to improve the original FCN, greatly improving the segmentation performance and boundary localization of the specific organs or diseased regions. SegNet~\cite{wickstrom2018uncertainty} is similar to UNet, but utilizes the max pooling indices to achieve up-sample operation in the decoder branch. SFANet~\cite{fang2019selective} incorporates a sharing encoder branch and two decoder branches to detect polyp regions and boundaries respectively, and includes a new boundary-sensitive loss to mutually improve both polyp region segmentation and boundary detection. In addition, by adopting the upward concatenation to fuse multi-level features and embedding the selective kernel module to learn multi-scale features, the model is further enhanced and achieves competitive results. However, most of the methods have not taken proper measures to deal with the shape and size variance of polyps regions.

In this paper, we propose the Adaptive Context Selection Network (ACSNet). Inspired by~\cite{fu2019adaptive}, we believe that the global context features are helpful for the segmentation of large polyps, while the local context information is crucial for the identification of small ones. Therefore, the intent of our designed network is to adaptively select context information as contrast learning and feature enhancement based on the size of the polyp region to be segmented. Specifically, our ACSNet is based on the encoder-decoder framework, with Local Context Attention (LCA) module, Global Context Module (GCM), and Adaptive Selection Module (ASM). LCAs and GCM are responsible for mining local and global context features and sending them to the ASM modules in each decoder layer. Through channel-wise attention, ASM well achieves adaptive feature fusion and selection. In summary, the contributions of this paper mainly include: (1) Our designed ACSNet can adaptively attend to different context information to better cope with the impact of the diversity of polyp size and shape on segmentation. (2) Our tailored LCA and GCM modules can achieve more consistent and accurate polyp segmentation through complementary selection of local features and cross-layer enhancement of global context. (3) ACSNet achieves new state-of-the-art results on two widely used public benchmark datasets.
\section{Method}
\begin{figure}
\centering
\includegraphics[width=0.95\textwidth]{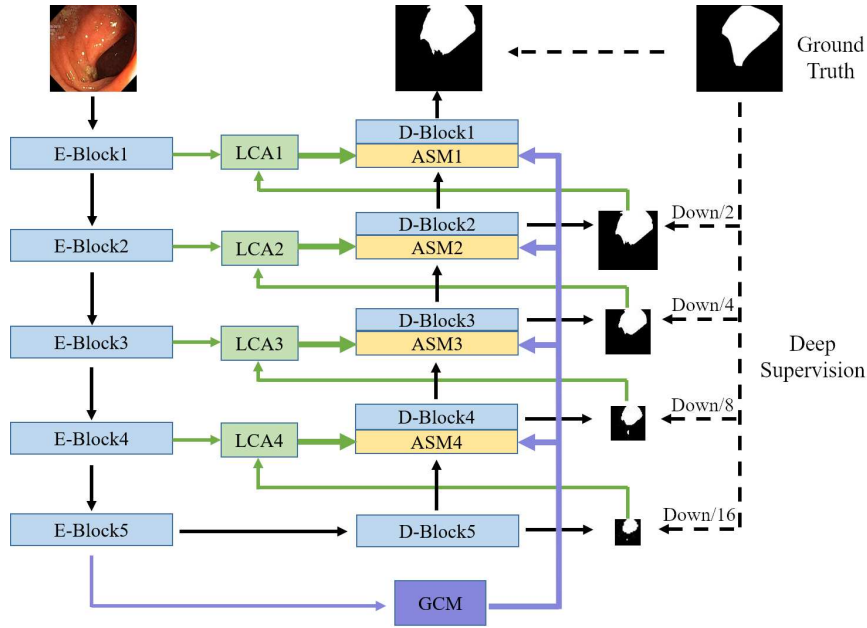}
\caption{Overview of our proposed ACSNet} \label{fig2}
\end{figure}

The architecture of our ACSNet is shown in Fig.~\ref{fig2}, which can be regarded as an enhanced UNet~\cite{ronneberger2015u} or Feature Pyramid Network (FPN)~\cite{lin2017feature}. We utilize ResNet34~\cite{he2016deep} as our encoder, which contains five blocks in total. Accordingly, the decoder branch also has five blocks. Each decoder block is composed of two Conv-BN-ReLU combinations, and generates one prediction map with different resolution, which is supervised by the down-sampled ground truth respectively.

The GCM is placed on top of the encoder branch, which captures the global context information and densely concatenates to the ASM of each layer in the decoder path. At the same time, each skip-connection between the encoder and decoder paths of UNet~\cite{ronneberger2015u} is replaced by the LCA module, which gives each positional feature column of every decoding layer a local context enhancement of different receptive field and at the same time delicately leverages the prediction confidence of the previous layer as a guidance to force the current layer to focus on harder regions. Finally, we utilize the ASM modules to integrate the features output from each previous decoder block, the LCA module and the GCM, based on a channel-wise attention scheme for context selection. 
\subsection{Local Context Attention Module (LCA)}
\begin{figure}[htp]
\includegraphics[width=0.8\textwidth]{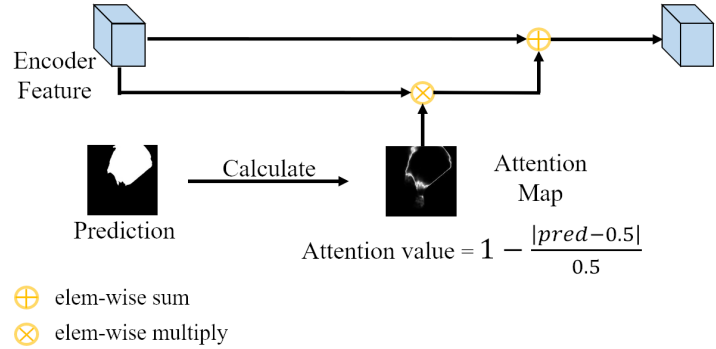}
\centering
\caption{Local Context Attention Module (LCA)} \label{fig3}
\end{figure}

LCA is designed as a kind of spatial attention scheme, which aims to incorporate hard sample mining when merging shallow features and pay more attention to the uncertain and more complex area to achieve layer-wise feature complementation and prediction refinement. As shown in Fig.~\ref{fig3}, the attention map of each LCA module is determined by the prediction map generated from the upper layer of the decoder stream. Specifically, the attention map of the $i^{th}$ LCA module is denoted as ${Att}_i\in\mathbb{R}^{1\times H_i\times W_i}$, in which $H_i$,$W_i$ are the height and width of the attention map respectively. The value of position $j\in[1,2,\cdots,H_i\times W_i ]$, denoted as ${Att}_{i}^{j}$ can be calculated as follows:
\begin{equation}
{Att}_{i}^{j}=1-\frac{\left|p^j_{i+1}-T\right|}{\max{(T,1-T)}}\ ,
\label{equ1}
\end{equation}
where ${P}_{i+1}^{j}\in(0,1)$ is the $j^{th}$ location value of the prediction map ${P}_{i+1}\in\mathbb{R}^{1\times H_i\times W_i}$ which is generated by the ${(i+1)}^{th}$ decoder block. $T$ is the threshold to determine whether the specific position belongs to foreground or background. We calculate the absolute difference between the prediction value and threshold T, and limit it to the range of 0 to 1 by dividing the maximum difference. We believe that the closer the predicted value is to the threshold T, the more uncertain the prediction of the corresponding position is, so it should be given a larger attention weight in the forwarding layer, and vice versa. Finally, we multiply the features by the attention values, and then sum with the original features to get the output of this module. For simplicity, $T$ is set to $0.5$ in our experiments.
\subsection{Global Context Module (GCM)}
We borrow the idea from pyramid pooling~\cite{zhao2017pyramid,liu2019simple,he2019non} to design our GCM and also put it as an independent module for global context inferring on top of the encoder branch. Meanwhile, GCM forwards the output to each ASM module to compensate the global context which is gradually diluted during layer-wise refinement. 

\begin{figure}[htp]
\centering
\includegraphics[width=0.8\textwidth]{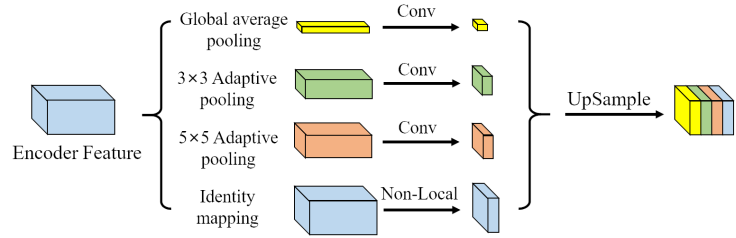}
\caption{Global Context Module (GCM)} \label{fig4}
\end{figure}

As shown in Fig.~\ref{fig4}, GCM contains four branches to extract context features at different scales. Specifically, this module is composed of a global average pooling branch, two adaptive local average pooling branches, and outputs three feature maps of spatial size $1\times1$, $3\times3$, $5\times5$, respectively. It also contains an identity mapping branch with non local operation~\cite{wang2018non} to capture the long range dependency while maintaining the original resolution. We introduce a non-local operation based feature representation here to finely capture the global dependency of each positional feature to enhance the output of the encoder. In the end, we up-sample the above four feature maps and concatenate them to obtain the resulted global context feature of this module, which will be densely fed to each designed ASM module in the decoder stream.
\subsection{Adaptive Selection Module (ASM)}
We believe that local context and global context have different reference values for the segmentation of polyp regions with different appearances, sizes, and feature contrasts. Therefore, we attach an adaptive context selection module (ASM) to each block in the decoder stream. Based on the local context features generated by the LCA, the global context features from the GCM, and the output features of previous decoder block as inputs, it learns to adaptively select context feature for aggregation in each block. 

\begin{figure}
\includegraphics[width=0.8\textwidth]{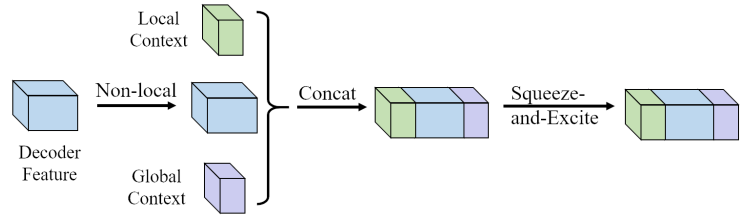}
\centering
\caption{Adaptive Selection Module (ASM)} \label{fig5}
\end{figure}

As shown in Fig.~\ref{fig5}, we incorporate a ``Squeeze-and-Excitation" block~\cite{hu2018squeeze} to adaptively recalibrate channel-wise feature responses for feature selection. Specifically, ASM takes the concatenated feature as input, and employs global average pooling to squeeze the feature map to a single vector which is further fed to a fully connected layer to learn the weight of each channel. After sigmoid operation, the attention weight is limited to the range of 0 to 1. Through multiplying the original feature maps with the attention values, some informative context features can be picked out while those not conducive to improving discrimination will be suppressed. Noted that we also apply non local operation~\cite{wang2018non} to the features output from previous decoder block before concatenation to enhance the decoder features with long range dependency.
\section{Experiments}
\subsection{Datasets}
We evaluate our proposed method on two benchmark colonoscopy image datasets, collected from the examination of colorectal cancer. The first is the EndoScene Dataset~\cite{vazquez2017benchmark}, which contains 912 images and each of which has at least one polyp region. It is divided into the training set, validation set and test set, with 547, 183, and 182 images respectively. For simplicity, we resize the images to $384\times288$ uniformly in our experiments. The second is Kvasir-SEG Dataset~\cite{jha2020kvasir} containing 1000 images with polyp regions. We randomly use 60$\%$ of the dataset as training set, 20$\%$ as validation set, and the remaining 20$\%$ as test set. Since the image resolution of this dataset varies greatly, we refer to the setting of~\cite{jha2020kvasir} and set all images to a fixed size of $320\times320$.  
\subsection{Implementation Details and Evaluation Metrics}
In the training stage, we use data augmentation to enlarge the training set, including random horizontal and vertical flips, rotation, zoom and shift. All the images are randomly cropped to $224\times224$ as input. We set batch size to 4, and use SGD optimizer with a momentum of 0.9 and a weight decay of 0.0005 to optimize the model. A poly learning rate policy is adopted to adjust the initial learning rate, which is $lr=init\_lr\times(1-\frac{epoch}{nEpoch})^{power}$, where $init\_lr=0.001$, $power=0.9$, $nEpoch=150$. We utilize the combination of a binary cross entropy loss and a dice loss as the loss function. Our model is implemented using PyTorch~\cite{paszke2019pytorch} framework.

As in~\cite{fang2019selective}, we use eight metrics to evaluate the segmentation performance, including ``Recall'', ``Specificity'', ``Precision'', ``Dice Score'', ``Intersection-over-Union for Polyp (IoUp)'', ``IoU for Background (IoUb)'', ``Mean IoU (mIoU)'' and ``Accuracy''.

\begin{figure}[t]
\centering
\includegraphics[width=\textwidth]{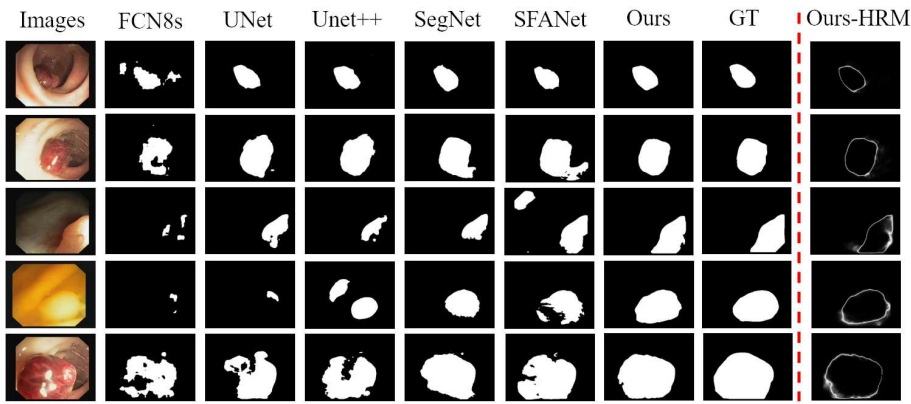}
\caption{Visual comparison of polyp region segmentation from state-of-the-art methods. The ground truth (GT) is shown in the penultimate column. Our proposed method consistently produces segmentation results closest to the ground truth. The hard region mining result is shown in the rightmost column.} \label{fig6}
\end{figure}

\begin{table*}[h]
\centering
\caption{Comparison with other state-of-the-art methods on the EndoScene dataset}
\resizebox{\textwidth}{!}{
\begin{tabular}{p{80pt}|p{30pt}|p{30pt}|p{30pt}|p{30pt}|p{30pt}|p{30pt}|p{30pt}|p{30pt}} 
\toprule
Methods & $Rec$  & $Spec$ & $Prec$ & $Dice$ & $IoUp$ & $IoUb$ & $mIoU$ & $Acc$ \\
\midrule
FCN8s~\cite{akbari2018polyp} &60.21&98.60&79.59&61.23&48.38&93.45&70.92&93.77 \\
UNet~\cite{ronneberger2015u}  &85.54&98.75&83.56&80.31&70.68&95.90&83.29&96.25 \\
UNet++~\cite{zhou2018unet++} &78.90&99.15&86.17&77.38&68.00&95.48&81.74&95.78 \\
SegNet~\cite{wickstrom2018uncertainty} &86.48&99.04&86.54&82.67&74.41&96.33&85.37&96.62 \\
SFANet~\cite{fang2019selective} &85.51&98.94&86.81&82.93&75.00&96.33&85.66&96.61 \\
\textbf{Ours}  &\textbf{87.96}&\textbf{99.16}&\textbf{90.99}&\textbf{86.59}&\textbf{79.73}&\textbf{96.86}&\textbf{88.29}&\textbf{97.11} \\
\bottomrule
\end{tabular}}
\label{table1}
\end{table*}

\subsection{Results on the EndoScene Dataset}
We compare our ACSNet with FCN8s~\cite{akbari2018polyp}, UNet~\cite{ronneberger2015u}, UNet++~\cite{zhou2018unet++}, SegNet~\cite{wickstrom2018uncertainty} and SFANet~\cite{fang2019selective} on the test set. As shown in Table.~\ref{table1}, our method achieves the best performance over all metrics, with $Dice$ of $86.59\%$, a $3.66\%$ improvement over the second best algorithm. Some visualization results are shown in Fig.~\ref{fig6} (Col.1-8), as can be seen that our algorithm is very robust to some complex situations such as polyp region sizes and image brightness changes. At the same time, due to the introduction of the effective context selection module and especially the hard region mining~(abbr.HRM) mechanism, the algorithm is significantly more accurate for polyp boundary positioning. In the rightmost column of Fig.~\ref{fig6}, it can be observed that the hard regions mined by our method are usually located in the border area of polyps, which is worthy of attention during prediction refinement.
\subsection{Results on the Kvasir-SEG Dataset}
On this dataset, we compare our ACSNet with UNet~\cite{ronneberger2015u}, UNet++~\cite{zhou2018unet++}, SegNet~\cite{wickstrom2018uncertainty}, ResUNet~\cite{jha2020kvasir} and SFANet~\cite{fang2019selective}. The results are listed in Table.~\ref{table2}. Similarly, our method achieves the best performance and outperforms others by large margins, further demonstrating the robustness and effectiveness of our method.
\begin{table*}
\centering
\caption{Comparison with other state-of-the-art methods and Ablation study on the Kvasir-SEG dataset}
\resizebox{\textwidth}{!}{
\begin{tabular}{p{100pt}|p{30pt}|p{30pt}|p{30pt}|p{30pt}|p{30pt}|p{30pt}|p{30pt}|p{30pt}} 
\toprule
Methods & $Rec$  & $Spec$ & $Prec$ & $Dice$ & $IoUp$ & $IoUb$ & $mIoU$ & $Acc$ \\
\midrule
UNet~\cite{ronneberger2015u}&87.89&97.69&83.89&82.85&73.95&94.73&84.34&95.65 \\
UNet++~\cite{zhou2018unet++} &88.67&97.49&83.17&82.80&73.74&94.49&84.11&95.42 \\
ResUNet~\cite{jha2020kvasir}&81.25&98.31&87.88&81.14&72.23&94.00&83.11&94.90 \\
SFANet~\cite{fang2019selective} &91.99&97.05&82.95&84.68&77.06&94.83&85.94&95.71 \\
SegNet~\cite{wickstrom2018uncertainty} &90.03&98.13&87.51&86.43&79.11&95.90&87.51&96.68 \\
\hline
\textbf{Ours}  &\textbf{93.14}&98.55&\textbf{91.59}&\textbf{91.30}&\textbf{85.80}&\textbf{97.00}&\textbf{91.40}&\textbf{97.64} \\
\hline
Baseline&89.53&98.63&90.32&88.21&81.59&96.27&88.93&96.99\\
Baseline+LCAs&91.79&98.39&89.15&89.00&82.47&96.41&89.44&97.15\\
Baseline+LCAs+GCM&92.18&\textbf{98.72}&90.90&90.28&84.35&96.88&90.62&97.52\\
\bottomrule
\end{tabular}}
\label{table2}
\end{table*}
\subsection{Ablation study}
To validate the effectiveness and necessity of each of the three modules in our proposed method, we compare ACSNet with its three variants in Table.~\ref{table2}. Specifically, the baseline model refers to the original U-shape encoder-decoder framework with skip-connections, and we gradually add LCAs, GCM, and  ASMs to it, denoted as Baseline+LCAs, Baseline+LCAs+GCM and Ours, respectively. As shown in the table, with the progressive introduction of LCAs, GCM, and ASMs, our algorithm has witnessed a certain degree of performance improvement, boosting $Dice$ by $0.79\%$, $1.28\%$, $1.02\%$ respectively.
\section{Conclusion}
In this paper, we believe that an efficient perception of local and global context is essential to improve the performance of polyps region localization and segmentation. Based on this, we propose an adaptive context selection 
based encoder-decoder framework which contains the LCA module for hard region mining based local context extraction, the GCM module for global feature representation and enhancement in each decoder block, and the ASM component for contextual information aggregation and selection. Extensive experimental results and ablation studies have demonstrated the effectiveness and superiority of the proposed method. 
\section*{Acknowledgement. }This work is supported in part by the Guangdong Basic and Applied Basic Research Foundation (No.2020B1515020048), in part by the National Natural Science Foundation of China (No.61976250 and No.61702565), in part by the ZheJiang Province Key Research $\&$ Development Program (No. 2020C03073) and in part by the Key Area R\&D Program of Guangdong Province (No. 2018B030338001).
%
%
%
\bibliographystyle{splncs04}
\bibliography{paper881}
\end{document}